\def\year{2022}\relax
\documentclass[letterpaper]{article} 
\usepackage{aaai22}  
\usepackage{times}  
\usepackage{helvet}  
\usepackage{courier}  
\usepackage[hyphens]{url}  
\usepackage{graphicx} 
\usepackage{natbib}  
\usepackage{caption} 
\DeclareCaptionStyle{ruled}{labelfont=normalfont,labelsep=colon,strut=off} 
\usepackage{algorithm}
\usepackage{algorithmic}

\usepackage{soul}
\usepackage{url}
\usepackage{multirow}
\usepackage{algorithm}
\usepackage{algorithmic}
\usepackage{epsfig}
\usepackage{amsmath}
\usepackage{graphics}
\usepackage{pifont}
\usepackage{amssymb}
\usepackage{newfloat}
\usepackage{listings}
\usepackage{color}
\definecolor{citecolor}{RGB}{0,0,0}
\definecolor{linkcolor}{RGB}{0,0,0}
\usepackage[table,xcdraw,dvipsnames]{xcolor}
\usepackage{hyperref}
\hypersetup{colorlinks=true,citecolor=citecolor,linkcolor=linkcolor,urlcolor=Thistle}
\usepackage{xspace}

\floatstyle{ruled}
\newfloat{listing}{tb}{lst}{}
\floatname{listing}{Listing}

\usepackage{fontenc}
\newcommand{\bt}{\color{black}}
\usepackage{bibentry}

\title{
Visual Semantics Allow for Textual Reasoning Better in Scene Text Recognition
}

\author{
    Yue He\textsuperscript{\rm 1},
    Chen Chen\textsuperscript{\rm 2},
    Jing Zhang\textsuperscript{\rm 2},
    Juhua Liu\textsuperscript{\rm 3}\footnote{Corresponding author. This work was done during Yue He's internship at JD Explore Academy.},
    Fengxiang He\textsuperscript{\rm 4},
    Chaoyue Wang\textsuperscript{\rm 2},
    Bo Du\textsuperscript{\rm 1}\footnotemark[1]
}

\affiliations{
    \textsuperscript{\rm 1} National Engineering Research Center for Multimedia Software, Institute of Artificial Intelligence, School of Computer Science and Hubei Key Laboratory of Multimedia and Network Communication Engineering, Wuhan University, China \\
    \textsuperscript{\rm 2} School of Computer Science, Faculty of Engineering, The University of Sydney, Australia\\
    \textsuperscript{\rm 3} School of Printing and Packaging, and Institute of Artificial Intelligence, Wuhan University, China \\
    \textsuperscript{\rm 4} JD Explore Academy, China\\
    \{yuehe.cs, liujuhua, dubo\}@whu.edu.cn, cche9000@uni.sydney.edu.au, \{jing.zhang1, chaoyue.wang\}@sydney.edu.au, hefengxiang@jd.com
}

\begin{document}
\maketitle

\begin{abstract}

Existing Scene Text Recognition (STR) methods typically use a language model to optimize the joint probability of the 1D character sequence predicted by a visual recognition (VR) model, which ignore the 2D spatial context of visual semantics within and between character instances, making them not generalize well to arbitrary shape scene text. To address this issue, we make the first attempt to perform textual reasoning based on visual semantics in this paper. Technically, given the character segmentation maps predicted by a VR model, we construct a subgraph for each instance, where nodes represent the pixels in it and edges are added between nodes based on their spatial similarity. Then, these subgraphs are sequentially connected by their root nodes and merged into a complete graph. Based on this graph, we devise a graph convolutional network for textual reasoning (GTR) by supervising it with a cross-entropy loss. GTR can be easily plugged in representative STR models to improve their performance owing to better textual reasoning. Specifically, we construct our model, namely S-GTR, by paralleling GTR to the language model in a segmentation-based STR baseline,
which can effectively exploit the visual-linguistic complementarity via mutual learning. S-GTR sets new state-of-the-art on six challenging STR benchmarks and generalizes well to multi-linguistic datasets. Code is available at \url{https://github.com/adeline-cs/GTR}.

\end{abstract}

\section{Introduction}

Scene Text Recognition (STR) remains a fundamental and active research topic in computer vision for its wide applications \cite{zhang2020empowering}. However, this task remains challenging in real-world deployment, since the recognition results are highly influenced by various factors, such as complex background, irregular shapes, diverse textures.

Existing methods mainly treat STR as a visual recognition (VR) task and perform character-level recognition on input images, including visual-text sequence translation-based methods{ \bt \cite{yang2017learning,shi2018aster,baek2019wrong,li2019show,litman2020scatter}} and semantic segmentation-based methods \cite{liao2019scene,wan2020textscanner}. Although these methods obtain reasonable performance on identifying individual characters, they ignore vital global textual representations, making it extremely hard to give robust recognition outcomes in real-world scenarios. 

\begin{figure}[t]
	\centering
	\includegraphics[width=\linewidth]{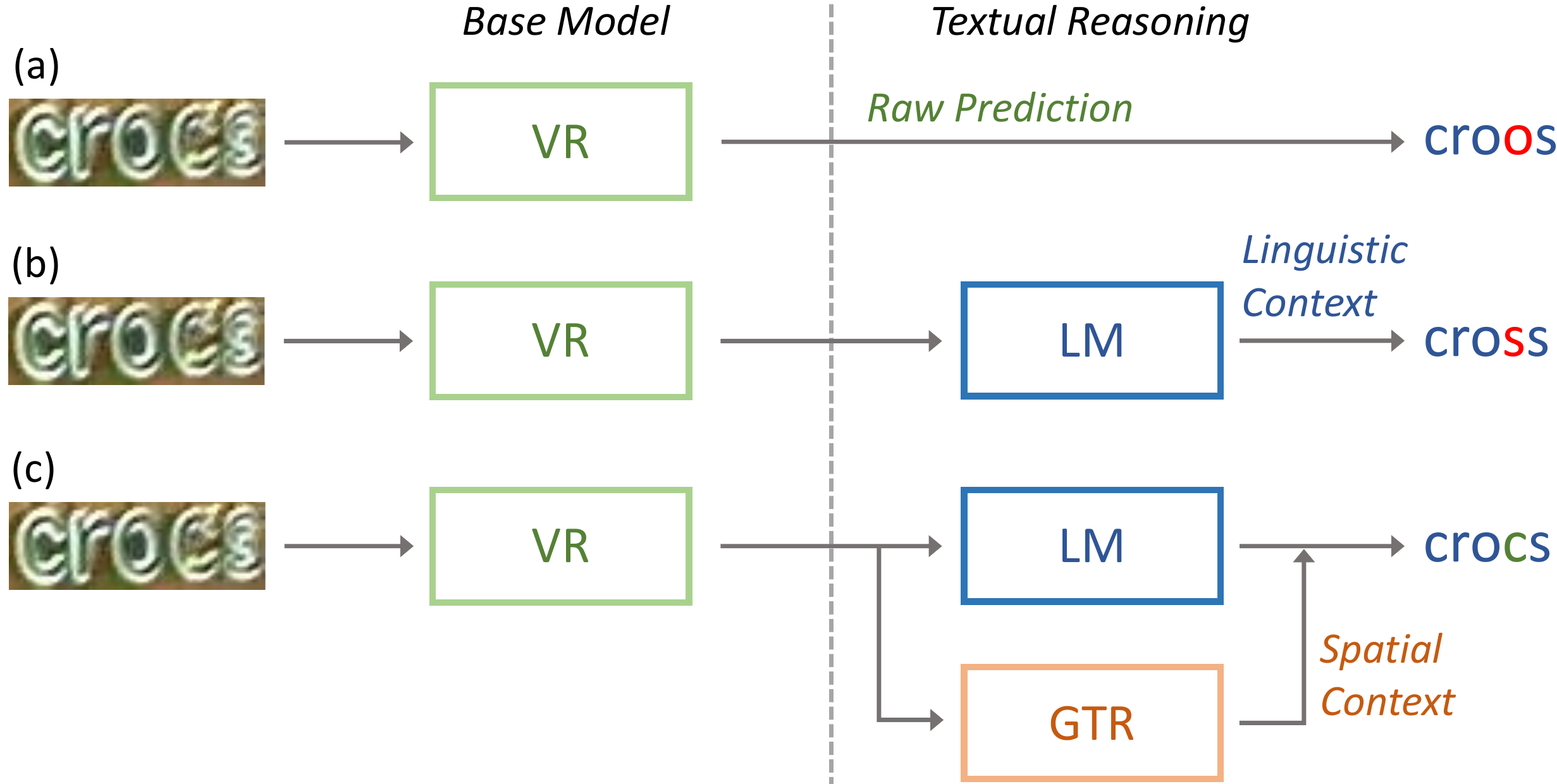}
	\caption{Diagram of different STR pipelines. (a) The VR model. (b) The VR model with an LM. (c) The proposed pipeline by adding GTR in parallel with LM. GTR performs textual reasoning based on the visual semantics generated by VR to address the irregular and blurry text. The ground truth label is ``crocs'' and wrong predictions are marked in red. }

	\label{fig:motivation}
	\vspace{-5mm}
\end{figure}

For global textual modeling, existing efforts \cite{qiao2020seed,yu2020towards,fang2021read} have been made to leverage a language model (LM) \cite{jaderberg2014deep} to optimize the joint probability of the character sequence predicted by the VR model. Though this strategy can correct mistaken predictions with linguistic context, it is hard to generalize to arbitrary texts and ambiguous cases. As shown in Fig. \ref{fig:motivation}(b), for the irregular and blurry text, even LM could not make correct predictions. Other than linguistic cues, spatial context could also contribute to global textual modeling of character sequences but few methods explore in this direction. Hence, existing models have difficulty in producing satisfactory results on irregular texts in diverse fonts and shapes as well as with blur and occlusions.

In this paper, we fill this gap with a novel \textbf{G}raph-based \textbf{T}extual \textbf{R}easoning (GTR) model for introducing spatial context into the text reasoning stage.
Given the character instances recognized by the VR model as well as the derived order relations between them, we first set up a two-level graph to establish the local-to-global dependency. In the first level, we construct a subgraph for pixels within each character instance based on their spatial similarity. And for the second level, $1$-st level subgraphs are merged into a complete graph by linking their root nodes, which represent the geometric center of all nodes within each subgraph. Accordingly, we devise a graph convolutional neural network for context reasoning and optimizing the joint probability of the character sequence.

Our proposed GTR is an easy-to-plug-in module and can seamlessly function with other context modalities. Specifically, we put GTR parallel to the LM to produce joint features for text reasoning (as shown in Fig. \ref{fig:motivation}(c)). To produce high-quality cross-modality representations, we design a mutual learning protocol to enforce the consistency between predictions from LM and GTR and employ a dynamic fusion strategy \cite{yue2020robustscanner} to deeply combine visual and linguistic features. Based on these designs, GTR can largely boost the text reasoning performance comparing to existing representative methods with LM only.

We incorporate all aforementioned designs into a segmentation-based STR baseline and propose S-GTR, a unified framework of \textbf{S}egmentation baseline with \textbf{GTR}. We evaluate S-GTR on multiple datasets with both regular and irregular text material in different languages. Experimental results show that our S-GTR outperforms previous methods and obtains state-of-the-art performance on six challenging benchmarks. In summary, the contribution of this work is 
threefold:

\begin{itemize}

\item We propose a novel graph-based textual reasoning model named GTR to refine coarse text sequence predictions with spatial context. It is a complementary design to the popular reasoning fashion with LM only in existing representative methods, and can further improve their performance by acting as an easy-to-plug-in module.

\item To make GTR work with LM compatibly, we further employ a mutual learning protocol and propose a dynamic fusion strategy to produce consistent linguistic and visual representations and high-quality joint prediction.

\item We put all our designs in a unified framework (S-GTR) of segmentation baseline with GTR. Extensive experimental results indicate our S-GTR successfully sets new state-of-the-art for regular and irregular text recognition tasks as well as shows a superiority on both English and Chinese text materials.

\end{itemize}

\section{Related Work}

\subsubsection{Arbitrary-shaped Scene Text Recognition.} 
Existing STR methods for recognizing texts of arbitrary shapes can be divided into two main categories, \textit{i.e.}, rectification-based methods and segmentation-based methods. The former methods \cite{gao2018recurrent, yang2017learning, cheng2018aon} use the spatial transformer network \cite{jaderberg2015spatial} to normalize text images into the canonical shape, $i.e.$, horizontally aligned characters of uniform heights and widths. These methods, however, are limited by the pre-defined text transformation set and hard to generalize to real-world examples. The latter methods \cite{liao2019scene,wan2020textscanner} {\bt follow the common processing fashion in the text detection task \cite{ye2021i3cl} and } formulate the recognition task as an instance segmentation problem, where texts are explicitly modeled into instance masks. In this way, it can efficiently represent irregular texts in different fonts, scales, orientation, and shapes, as well as with occlusions. For this reason, we choose to build our GTR model upon a base instance segmentation-based recognition model. In addition, since the segmentation probability maps embed useful semantics and spatial context, the propose GTR model can efficiently exploit them for text reasoning.


\subsubsection{Semantic Context Reasoning.}
To further enhance the text recognition performance, some methods resort to linguistic context to improve raw outputs from the VR model. For example, \cite{cheng2017focusing}
employ {\bt CNN} to yield bags of N-grams of text string for output reasoning. 
Some later methods \cite{Wang2020DecoupledAN,wojna2017attention} leverage {\bt RNN} to strengthen context dependencies between characters.
Recently, some methods adopt semantic context reasoning to achieve high performance. SEED \cite{qiao2020seed} proposes to use word embedding from FastText \cite{bojanowski2017enriching}, which relies on semantic meaning of a word instead of its visual appearance.
SRN \cite{yu2020towards} uses transformer-based models where global semantic information as well as long-range word dependency is modelled by self-attention. It it computationally efficient due to the parallel nature of transformer architecture {\bt like \cite{xu2021vitae}}, but their non-differentiable semantic reasoning block imposes a significant limitation. 
{\bt Based on SRN, ABINet \cite{fang2021read} adopts the iterative correction for enhancing semantic reasoning.}  
Beyond semantic reasoning, we propose a graph-based context reasoning model that supplements the language model to exploit both visual spatial context and linguistic context to improve the visual recognition results.

\begin{figure*}[t]
	\centering
    \includegraphics[width=1.0\textwidth]{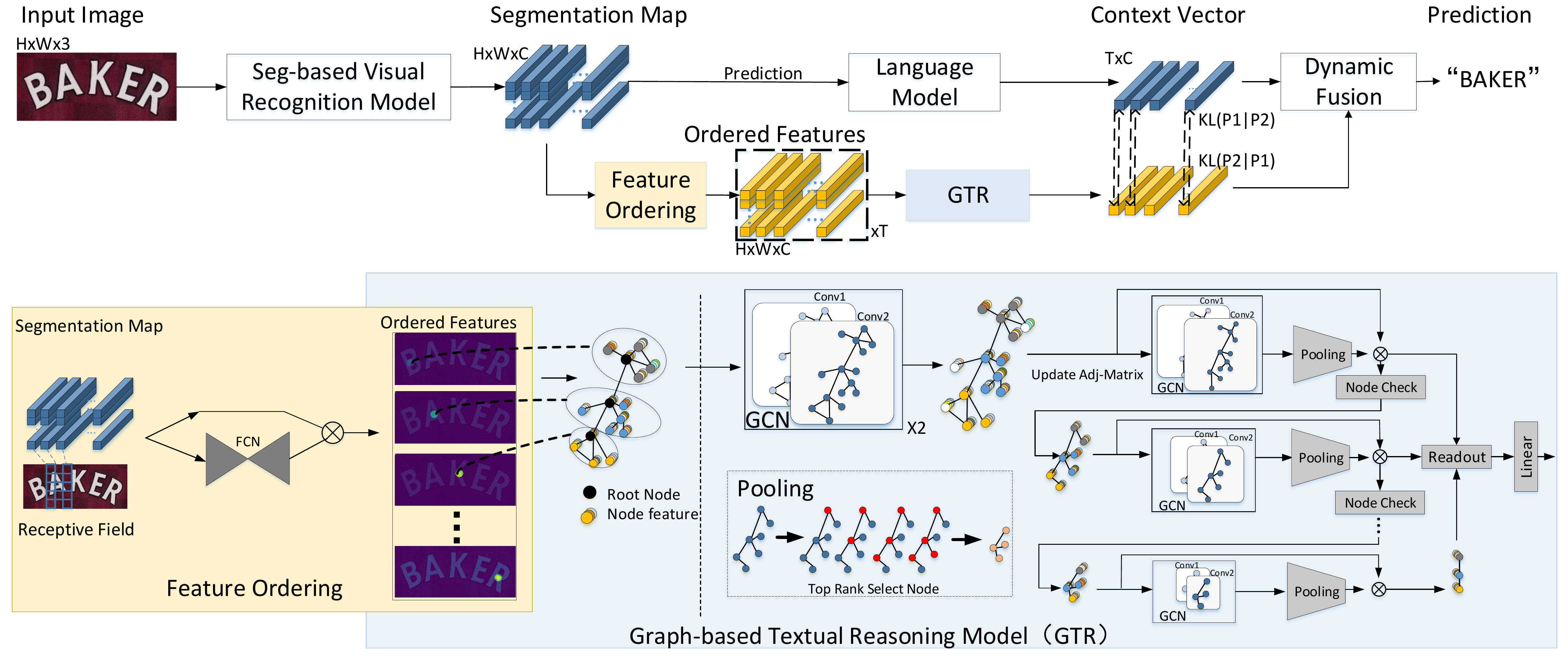} 
	\caption{Overview of the proposed S-GTR model. It consists of a VR model, a LM, and the proposed GTR. GTR is stacked on the top of the VR model and in parallel with the LM. The detailed structure of GTR as well as a pre-processing step, $i.e.$, feature ordering, are also shown in the bottom part of this figure. More details can be found in Section~\ref{sec:method}.}

	\label{fig:overview}
\end{figure*}

\subsubsection{Graph-structure Data Reasoning.}
\label{subsec:gcn}
Considerable efforts have been made to design graph convolutional neural networks (GCN) for modelling graph-structured data \cite{kipf2016semi,chen2019graph,wang2019linkage}. For example, 
in the text detection task, \cite{zhang2020deep} adopts a GCN to link characters that belong to the same word. GTC \cite{hu2020gtc} utilizes a GCN to guide CTC \cite{graves2006connectionist} for scene text recognition. Specifically, it models the sliced visual features as the graph nodes, captures their dependency, and merges features of the same instance for prediction. 
{ \bt PREN2D \cite{yan2021primitive} adopts a meta-learning framework to extract visual representations via GCN.}
In this paper, we devise a two-level graph network based on GCN to perform spatial context reasoning
within and between character instances to refine the visual recognition results.

\section{Methodology}
\label{sec:method}

\subsection{Overview}
The full framework of S-GTR is shown in the Figure \ref{fig:overview}, which comprises a segmentation-based VR model, an LM, and our proposed GTR. Given the input image $\mathbf{X} \in \mathbb{R}^{H\times W \times 3}$, the segmentation-based VR first produces a segmentation map $\mathbf{M} \in \mathbb{R}^{H\times W \times C}$, where $C$ is the number of character classes. The segmentation map $\mathbf{M}$ is decoded to a preliminary text sequence prediction $\mathbf{T} \in \mathbb{R}^{T\times C}$ and further processed by LM for generating linguistic context vector $\mathbf{L} \in \mathbb{R}^{T\times C}$. $T$ is the pre-defined maximum length of output sequence. 

Our proposed GTR is stacked in parallel with LM, taking the segmentation map $\mathbf{M}$ as input. Firstly, we transform the map $\mathbf{M}$ with a feature ordering module to build an ordered feature tensor $\mathbf{V} \in \mathbb{R}^{T\times H\times W\times C}$, which comprises $T$ attention maps and represents the relations between geometry features and text order information. Next, we build a sub-graph for each attention map and then connect all sub-graphs sequentially into a full graph. The graph is then deeply encoded with a GCN to produce the spatial context vector $\mathbf{S} \in \mathbb{R}^{T\times C}$. Finally, the coarse sequence prediction $\mathbf{T}$, the linguistic context $\mathbf{L}$ and the spatial context $\mathbf{S}$ are combined via dynamic fusion and the refined text is predicted.

\subsection{GTR: Graph-based Textual Reasoning}
\label{subsec:gtr}
Given the segmentation map $\mathbf{M}$, we employ a fully convolutional network (FCN) to obtain a series of attention maps related to the character order and use them to attend $\mathbf{M}$ via element-wise multiplication to get the ordered feature tensor $\mathbf{V} \in \mathbb{R}^{T\times H\times W\times C}$, as shown in the bottom left part of Figure~\ref{fig:overview}. Based on $\mathbf{V}$, GTR firstly builds sub-graphs for all character instances and connects them sequentially. Then, the graph is encoded with a GCN and pooling operation to produce spatial context.

\begin{figure}[t]
	\centering
	\includegraphics[width=1.0\columnwidth]{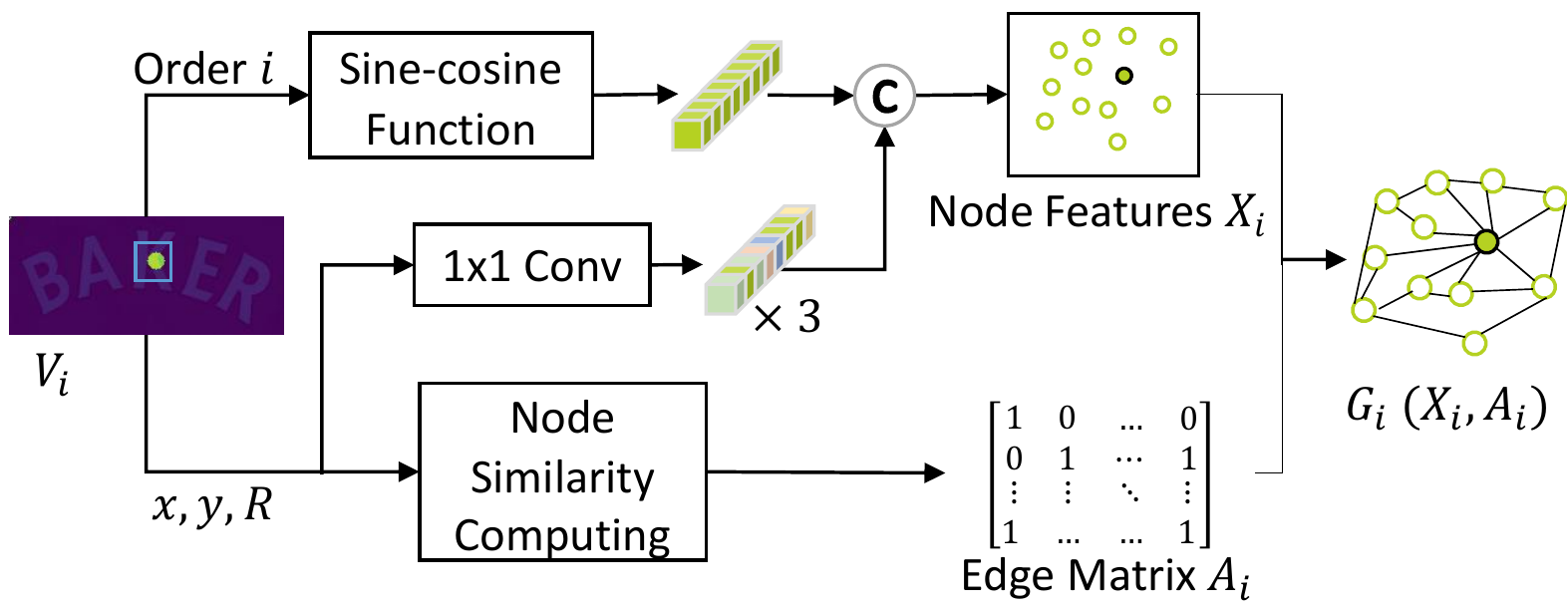}
    \caption{Illustration of the subgraph building process, including node feature generation and edge matrix compute.}
	\label{fig:graph}
\end{figure}

\subsubsection{Graph Generation}

We build the two-level graph from the ordered feature tensor $\mathbf{V}$ to model the local-to-global dependency. We first connect pixels belonging to the same character in the $1$-st level sub-graph. Specifically, for the $i$-th ordered feature map $V_i \in \mathbb{R}^{H\times W\times C}$, we first choose pixels having the same estimation to the $i$-th character in the text sequence predicted by VR. These pixels are collected as a set $P_i = \{(x,y,R)_j\}$, where $R$ is the $C$-dim feature vector in the position $(x, y)$ of $V_i$ and $j$ is the pixel index. Note that we also add a root node with average $x$, $y$ and $R$ to the set. Then we construct the node feature vector $X_{i,j}$ by embedding $x$, $y$ and $R$ with three different $1\times 1$ convolutions and $i$ with sine and cosine functions. The four parts of embeddings are concatenated to form node features.

Next, the adjacent matrix is built according to node similarities. We compute the node similarity as a product between the position similarity $E_p$ and the feature similarity $E_f$, which is defined as:
\begin{equation}
	E_{p}(p,q) = 1-\frac{D\left ( (x_p, y_p) , (x_q, y_q)\right )}{max\left (H ,W \right )},
\end{equation}
\begin{equation}
    E_{f}(p,q) = \frac{R_{p}\cdot R_{q}}{\left \| R_{p}, \right \| \left \| R_{q} \right \|},
\end{equation}
\begin{equation}
    E(p,q) = E_{p}(p,q) \cdot E_{f}(p,q),
    \label{eq:ps}
\end{equation}
where $p$ and $q$ are two nodes from the set $P_i$. The position similarity $E_p$ is negatively proportional to the Euclidean distance between two pixels whereas the feature similarity $E_f$ is the cosine similarity between pixel features. The product of $E_p(p,q)$ and $E_f(p,q)$ is the overall similarity $E$ between node $p$ and $q$. Then, we use the 1-hop rule \cite{wang2019linkage} to build the adjacent matrix $A_i$. Specifically, we connect each node in $V_i$ to other nodes that have the top-$8$ largest similarity and delete the connections to the nodes outside the 1-hop cluster.

After constructing sub-graphs $G_i(X_i, A_i)$, we connect them into the $2$-level full graph by linking their root nodes in sequence order. The full graph is denoted as $G(X, A)$.

\subsubsection{Spatial Context Reasoning}

Given a graph $G(X,A)$, we try to use the graph convolutional network to perform two-stage spatial context reasoning by following \cite{zhang2020deep, kipf2016semi}.

The first stage is spatial reasoning. After obtaining the feature matrix $X$ and the adjacency matrix $A$, we use a graph convolutional network to output a transformed node feature matrix $Y$.
This process can be described as follows:
\begin{equation}
		Y^l = \sigma ([X^{l}; KX^{l}]W^{l}), l=1...L,
\end{equation}
\begin{equation}
		K=\tilde{D}^{\frac{-1}{2}}\tilde{A}\tilde{D}^{\frac{-1}{2}}.
		\label{eq:G}
\end{equation}
Here, $l$ denotes the layer index, $L=2$, $X(l)\in \mathbb R^{N\times d_{i}}$, $Y(l) \in \mathbb R^{N \times d_{o}} $, $d_{i}$ and $d_{o}$ are the dimension of input and output node features, and $N$ is the number of nodes. $[;]$ represents concatenation. $W^{l} $ is a layer-specific trainable weight matrix. $\sigma$ denotes a non-linear activation function. $K$ is an aggregation matrix of size $N \times N$, which is computed according to \cite{kipf2016semi}. Note that $X^{l+1}=Y^l$, $i.e.$, the output feature matrix $Y^l$ is used as the input of the $l+1$th layer.

After spatial reasoning, we perform the contextual reasoning. Denoting the output graph feature matrix from the aforementioned graph convolutional network as $X_c^l$, we compute a new adjacency matrix $A_{c}$ based on $X_c^l$. Then, we calculate $G$ according to Eq.~\eqref{eq:G} based on $A_{c}$. Next, we use a graph convolutional network to output a transformed node feature matrix $Y_c^l$ as follows:
\begin{equation}
	Y_{c}^l = \sigma ((GX_{c}^{l})W_c^{l}), l=1...L,
\end{equation}
where $W_c^{l}$ is a layer-specific trainable weight matrix.

Then, we perform root node check to make sure the selected root node is the underlying reliable root node, $i.e.$, the center of the character instance. In this way, it achieves the balance between the edges with near easy nodes and distant hard nodes by satisfying the following criterion:
\begin{equation}
	G_{iou}=\frac{G_{r}\cap G_{s}}{G_{r}\cup G_{s}} < \varepsilon,
\end{equation}
where $G_{r} $ and $G_{s} $ are two subgraphs for a same character, given that $s$ is a randomly selected node as the root node while $r$ is always the center of character. $G_{r}\cap G_{s}$ and $G_{r}\cup G_{s}$ are the intersection and the union of 1-hop neighbors of $G_{r} $ and $G_{s}$, respectively. In our experiments, $\varepsilon$ is set to 0.75.

Next, we use a readout layer like \cite{Lee2019SelfAttentionGP} to aggregate node features to a fixed-size representation. The output feature of this layer is calculated as follows:
\begin{equation}
	x_i^{*} = [x_{i};max({x_{j}^{*}|j\in N(x_i)})],
	\label{eq:readout}
\end{equation}
where $x_{j}^{*}$ is the updated feature at $j$th node, which is also calculated according to Eq.~\eqref{eq:readout}, $i.e.$, $x_i^{*}$ is calculated in a recursive manner. $N(x_i)$ denotes the neighboring node set of node $i$. 
After we obtain the updated node features, we discard 50\% nodes that are most distant to the root node, $i.e.$, pooling the graph into a smaller new one. We iteratively repeat the feature update and pooling process until only a node exists in the subgraph, resulting in a node sequence. Finally, the feature representations of the node sequence are passed to a linear layer for classification. We adopt the softmax cross-entropy loss for optimizing graph convolutional neural network. Similar to \cite{wang2019linkage}, we only back-propagate the gradient for nodes in the 1-hop neighborhood during training.

\begin{table*}[t]
	\footnotesize
	\centering
	\begin{tabular}{l|c|c|c|c|c|c|c|c|c}
		\hline
		\multirow{2}{*}{Methods} & \multirow{2}{*}{Training Data}   & \multicolumn{3}{c}{Regular} & \multicolumn{3}{|c|}{Irregular} & \multirow{1}{*}{Params} &\multirow{1}{*}{Time}  \\
		\cline{3-8}
		&  &IIIT5k & SVT  & IC13 & SVTP & IC15 &CUTE & ($\times 10^{6}$) & (ms)  \\
		\hline
		CRNN \cite{shi2016end} &ST + MJ & 78.2 & 80.9 & 89.4 & -	 &-	  & - & 8.3  & 6.8\\
		FAN* \cite{cheng2017focusing} & ST + MJ  & 87.4 & 85.9 & 93.3 & -  & 70.6 & -  & - & - \\ 
		ASTER \cite{shi2018aster}& ST + MJ  & 93.4 & 89.5 & 91.8 & 78.5 & 76.1 & 79.5 & 22 & 73.1\\
		CA-FCN* \cite{liao2019scene}& ST + MJ & 91.9 & 86.4 & 91.5 & -  & - & 79.9 & -&-\\
		TRBA \cite{baek2019wrong} &ST + MJ & 87.9 & 87.5 & 92.3 & 79.2 &77.6  & 74.0 &49.6 &27.6\\
		Textscanner* \cite{wan2020textscanner}  & ST + MJ &93.9	&90.1 &92.9 &84.3 &79.4 &83.3& 57   & 56.8\\  
		GTC \cite{hu2020gtc}&ST + MJ & 95.5 & 92.9 & 94.3 & 85.7 &  79.5 & 92.2 &- &-\\
		SCATTER \cite{litman2020scatter} &ST + MJ &93.2 &90.9 &94.1 &86.2 &82.0 &84.8 &- & - \\
		SEED \cite{qiao2020seed} &ST + MJ &93.8 &89.6 &92.8 &81.4 &80.0 &83.6&- &- \\
		SRN \cite{yu2020towards}& ST + MJ & 94.8 & 85.1 & 95.5 & 85.1 & 82.7 & 87.8 & 49.3 & 26.9\\
		RobustScanner \cite{yue2020robustscanner} &ST + MJ  &95.3&88.1&94.8& 79.5 &77.1 &90.3 &-&-\\
		Base2D \cite{yan2021primitive} & ST + MJ &95.4 &93.4 &95.9 &86.0 &81.9 & 89.9 & 59.0 &61.6 \\ 
		PREN2D \cite{yan2021primitive} &ST + MJ  &95.6 &94.0 &96.4 &87.6 &83.0 & 91.7 & - &67.4 \\
        \bt ABINet-LV$^\dag$ \cite{fang2021read}  &ST + MJ   &96.3   & 93.0  & 97.0  & 88. 5   & 85.0  & 89.2   & 36.7 & 22.0\\
		\hline
		Seg-Baseline &ST + MJ  &94.2	&90.8 &93.6	&84.3 &82.0 &87.6 & 34.0 & 14.0\\
		\textbf{S-GTR}         &ST + MJ  &\bf{95.8} & \bf{94.1} & \bf{96.8} &\bf{87.9}	&\bf{84.6}	& \bf{92.3} &42.1 & 18.8\\ 
		\textbf{GTR} + CRNN$^{[\text{CTC}]}$ &ST + MJ  & 87.6 & 82.1 & 90.1& 68.1 & 68.2 & 78.1 &15.2 &12.8\\
		\textbf{GTR} + TRBA$^{[\text{1DATT}]}$ &ST + MJ  & 93.2 & 90.1 & 94.0& 80.7 & 76.0 & 82.1 & 54.2 & 32.9\\
		\textbf{GTR} + SRN$^{[\text{Transformer}]}$ &ST + MJ   & 96.0 & 93.1 & 96.1 & 87.9 & 83.9 & 90.7 &54.3 &31.6\\
		\textbf{GTR} + Base2D$^{[\text{2DATT}]}$  &ST + MJ  & 96.1 & 94.1 & 96.6 & 88.0 & 85.3 & 92.6& 64.1 & 65.7\\
        \bt \textbf{GTR} + ABINet-LV$^{\dag[\text{Transformer}]}$ &ST + MJ & 96.8   & 94.8  & 97.7  & 89.6   & 86.9  & 93.1
        & 41.6 & 30.9 \\
		\hline 
		SAR\cite{li2019show}& ST + MJ + R & 95.0 & 91.2 & 94.0 & 86.4 & 78.0 & 89.6 &- &-\\
		Textscanner* \cite{wan2020textscanner}  & ST + MJ + R & 95.7 & 92.7 & 94.9 & 84.8 & 83.5 & 91.6 & 57&56.8\\ 
		RobustScanner \cite{yue2020robustscanner} &ST + MJ + R &95.4&89.3&94.1& 82.9&79.2&92.4 &- &-\\
		ABINet \cite{fang2021read} &ST + MJ + R  &97.2	&95.5 &97.7	&90.1 &86.9	&94.1 & -&-\\ 
		\bf{S-GTR}    &ST + MJ + R  &\bf{97.5}	&\bf{95.8}	&\bf{97.8}	&\bf{90.6}	&\bf{87.3}	&\bf{94.7} &42.1 &18.8 \\ 
		\hline
	\end{tabular}
	\caption{Results of our S-GTR, SOTA methods and their variants with our GTR on six regular and irregular STR datasets. ``R'' denotes the real datasets. ``*'' means using character-level annotations during training. {\bt ``\dag'' means the batch size is set to 384 for a fair comparison.} The superscripts in the second group of rows denote the type of different methods, $i.e.$, ``CTC'': CTC-based method, ``1DATT'': 1D attention-based method, ``2DATT'': 2D attention-based method, and ``Transformer'': Transformer-based method. Details can be found in Section~\ref{subsec:performanceAnalysis}.}
	\label{tab:main}
\end{table*}


\subsection{S-GTR: A Simple Baseline for STR}

We incorporate our GTR to a popular segmentation-based VR model with LM, resulting in a simple baseline for STR, $i.e.$, S-GTR, as shown in Figure~\ref{fig:overview}. 
{\bt Specifically, the VR model is designed following \cite{wan2020vocabulary}, and the LM is based on SRN.}
We devise manifold training strategies to make GTR better support the the STR task.

\subsubsection{Context Consistency}
Since we have two different types of reasoning features, namely linguistic context and spatial context. To prevent S-GTR from overly relying on one of them and avoid inconsistent reasoning cues to provide ambiguous results, we propose a mutual learning strategy to enforce the consistency between the two types of context features. Specifically, we compute the Kullback Leibler (KL) divergence between $\mathbf{L}$ from LM and $\mathbf{S}$ from GTR. 

\subsubsection{Dynamic Fusion}
\label{subsec:dynamicfusion}
Following \cite{yue2020robustscanner} that uses a dynamic fusion module to fuse information from multiple domains, we extend it in S-GTR to combine three different text sequences from VR, LM and GTR.Formally,
\begin{equation}
\begin{aligned}
   Q_{i} = Sigmoid(W_{0}[\mathbf{T}_i;\mathbf{L}_i;\mathbf{S}_i]), \\
    Z_{i} = Q_{i}\odot(W_{1}[\mathbf{T}_i;\mathbf{L}_i;\mathbf{S}_i]),
\end{aligned}
\end{equation}
where $\mathbf{T}_i$, $\mathbf{L}_i$, $\mathbf{S}_i$ are prediction vectors for the $i$-th character. $W_{0}$ and $W_{1}$ are two learnable linear transformations and $\odot$ indicates the element-wise multiplication operation. $Z_i$ is the final output of S-GTR for the $i$-th character.

\subsubsection{Mean Teacher-based Syn-to-Real Adaptation}
To mitigate the domain shift issue when using both synthetic and real datasets for training, we adopt the popular mean teacher framework \cite{meanteacher} in the area of domain adaptation. Specifically, a teacher network with the identical architecture as the segmentation-based VR model ($i.e.$, student network) is built and its weights are the exponential moving average of those of the student network.

\subsubsection{Loss Function}
The overall loss contains three parts, including sequence prediction loss $\mathcal{L}_{\text{Seq}}$, the LM-GTR consistency loss $\mathcal{L}_{\text {CC}}$, and the mean-teacher training loss $\mathcal{L}_{\text {MT}}$:
\begin{align} 
	L = \lambda_{\text{Seg}}*\mathcal{L}_{\text{Seg}}+\lambda_{\text{CC}}*\mathcal{L}_{\text{CC}}+\lambda_{\text{MT}}*\mathcal{L}_{\text {MT}}.
\end{align}
$\mathcal{L}_{Seg}$ contains a cross-entropy loss for character classification and a smooth L1 loss for order segmentation. $\mathcal{L}_{\text{CC}}$ is the KL loss for context consistency. $\mathcal{L}_{\text{MT}}$ is the MSE loss on the segmentation maps from teacher and student networks. $\lambda_{\text{Seg}}$ and $\lambda_{\text{CC}}$ are both set to 1.0. $\lambda_{\text{MT}}$ is set to 1.0 when using synthetic datasets for training. After getting accurate feature representations, it is reduced to $0$ gradually.

\section{Experiments}
\subsection{Experimental Settings}
\subsubsection{Datasets}
Following \cite{yu2020towards}, we use two public synthetic datasets , $i.e.$, SynthText (ST) \cite{gupta2016synthetic} and MJSynth (MJ) \cite{jaderberg2014synthetic,jaderberg2016reading} and a real datasets (R) \cite{baek2021if} for training. We test the trained model on six benchmarks including three regular scene-text datasets, $i.e.$, ICDAR2013 \cite{karatzas2013icdar}, IIIT5K \cite{mishra2012scene}, SVT \cite{wang2011end}, and three irregular text datasets, $i.e.$, ICDAR2015 \cite{karatzas2015icdar}, SVTP \cite{phan2013recognizing} and CUTE \cite{risnumawan2014robust}. The evaluation metric is the standard word accuracy.

\subsubsection{Implementation Details}
\label{subsec:arch-details}
We train the model with ADAM optimizer on two synthetic datasets for $6$ epochs and then transferred to the real dataset for the other $2$ epochs. The total batch size is $256$, equally distributed on four NVIDIA V100 GPUs. For the pre-training stage on synthetic datasets, the learning rate is set to $0.001$ and divided by $10$ at the $4$-th and $5$-th epochs. Then, we utilize the mean teacher training framework on the real dataset for the remaining $2$ epochs. The detailed training setting for this stage is deferred to the supplementary material.

Our model recognize $63$ types of characters, including ``$0$-$9$", ``a-z", and ``A-Z". The max decoding length of the output sequence $T$ is set to 25. We follow the standard image pre-processing that randomly resizing the width of original images into 4 scales, \textit{i.e.}, $64$, $128$, $192$ and $256$, and then padding the images to the resolution of $64 \times 256$. We adopt multiple data augmentation strategies including random rotation, perspective distortion, motion blur, and adding Gaussian noise to the image.

\subsection{Performance Analysis}
\label{subsec:performanceAnalysis}
\subsubsection{Comparison with State-of-the-Art}

We compare the proposed S-GTR with state-of-the-art methods, and the results are summarized in Table~\ref{tab:main}, where the inference speed as well as the number of model parameters are also reported. As can be seen, the proposed S-GTR achieves the highest recognition accuracy and $3\times$ faster inference speed compared with the second best method PREN2D \cite{yan2021primitive}. In addition, when real data is utilized for training, S-GTR achieves more impressive results on all the six benchmarks, validating the effectiveness of the proposed GTR for textual reasoning and the benefit of real data.

\subsubsection{Plugging GTR in Different Models}

To further verify the effectiveness of GTR, we plug our GTR module into four representative types of STR methods, including CTC-based method ($e.g.$, CRNN \cite{shi2016end}), 1D attention-based method ($e.g.$, TRBA \cite{baek2019wrong}), 2D attention-based method ($e.g.$, Base2D \cite{yan2021primitive}), and transformer-based methods ($e.g.$, SRN \cite{yu2020towards} and {\bt ABINet-LV} \cite{fang2021read}). For the 1D attention-based method, the prediction result of VR is a 1D semantic vector. Therefore, we adopt the 2D feature map from the layer before the prediction layer as input of GTR after feature ordering. The results are shown in the second group of rows in Table~\ref{tab:main}. As can be seen, after using GTR, the performance of all these models can be improved further. For example, the average recognition accuracy on all the available test sets is increased by $3.77\%,3.20\%,2.78\%,1.69\%$, and {\bt $1.65\%$} for CRNN, TRBA, SRN, Base2D, and {\bt ABINet-LV}, respectively. These results demonstrate the compatibility of our GTR to typical models.


\begin{figure}
	\centering
    \includegraphics[width=1.0\linewidth]{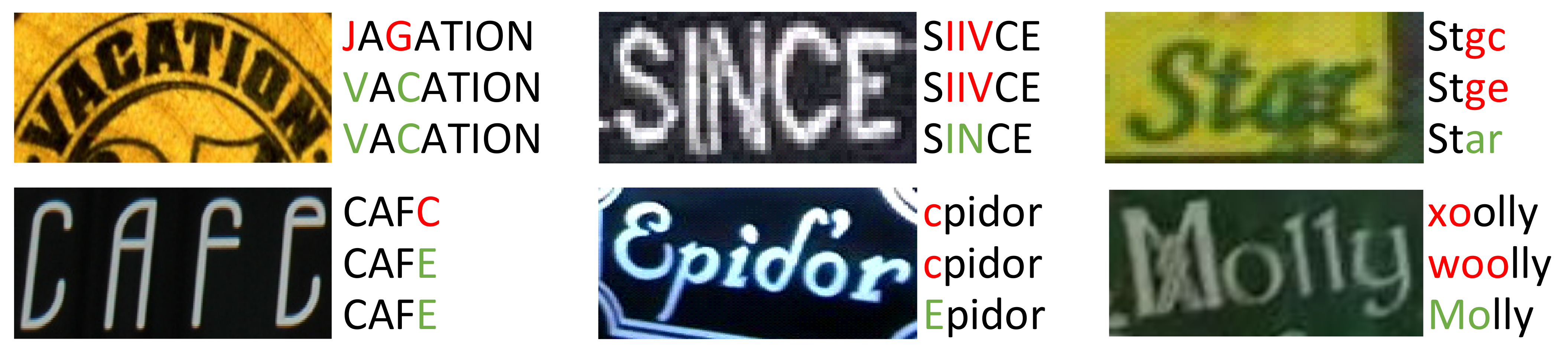}
	\caption{Results on some test images. Each image is along with three texts, which are predicted by VR, the Seg-baseline, and the proposed S-GTR model, respectively.}
	\label{main-pic}
\end{figure}
\subsection{Ablation Study}
\subsubsection{Ablation Study Results of S-GTR}
All the models in this ablation study have the same training configurations as used in S-GTR. To investigate the impact of different modules in S-GTR, we first train a baseline VR model, which utilizes neither LM and nor GTR. As shown in Table~\ref{tab:module}, without LM and GTR, the baseline VR model observes a significant performance drop. Compared with the baseline, a gain of $3.45\%$ can be observed by using LM, since it introduces the global linguistic textual cues for textual reasoning and corrects some linguistically implausible predictions. The proposed GTR module exploits the visual-spatial context information to refine the output of the VR model and increases the average accuracy by $4.06\%$. When using both LM and GTR together, the best average performance of $90.96\%$ can be obtained. These two modules both contribute to the improvement of S-GTR over the baseline, demonstrating that the linguistic cues and spatial context from visual semantics are complementary to each other. It is also noteworthy that the GTR module brings more gains than LM. 

Note that there is no use of mutual learning in the experiments in Table~\ref{tab:module}. After comparing the results in its last row ($i.e.$, S-GTR without mutual learning) with the results of S-GTR in Table~\ref{tab:main}, which is also trained on the ``ST+MJ'' datasets but with mutual learning, we can find that mutual learning contributes to a better average recognition accuracy of $91.92\%$. It demonstrates that enforcing the consistency between the context features from LM and GTR is necessary to better exploit the complementary between these two different types of textual reasoning.


For the qualitative analysis of different models, we present some test images and their corresponding text predictions from the basic VR model (top), Seg-baseline with LM (middle), and the proposed S-GTR (bottom) in Figure~\ref{main-pic}. We can see that LM can correct some mistaken predictions from the basic VR model by exploiting the global linguistic context. However, it is still challenging to generalize to arbitrary texts and some ambiguous cases. Compared to it, S-GTR produces satisfactory results on irregular texts in different fonts, scales, orientations, and shapes, owing to its better textual reasoning ability by exploiting both linguistic cues and spatial context from visual semantics.

\begin{table}[t]
\resizebox{1.0\linewidth}{!}{

\begin{tabular}{ccccccccc}
\hline
\multicolumn{2}{c}{Seg-baseline} & \multirow{2}{*}{GTR} & \multirow{2}{*}{IIIT5k} & \multirow{2}{*}{SVT}  & \multirow{2}{*}{IC13} & \multirow{2}{*}{SVTP} & \multirow{2}{*}{IC15} & \multirow{2}{*}{CUTE}\\ 
VR & LM \\
  \hline
\checkmark & & &91.8 &86.6 &91.1 &79.8 &77.7 &84.8 \\ 
\checkmark &\checkmark & &94.2 &90.8 &93.6 &84.3 &82.0 &87.6 \\ 
\checkmark & &\checkmark &94.0 &91.2 &94.8 &85.0 &82.8 &88.4 \\ 
\checkmark &\checkmark &\checkmark & 95.1 &93.2 &95.9 & 86.2 &84.1 &91.3 \\ 
\hline
\end{tabular}
}
\caption{Ablation study of the components in S-GTR.}
\label{tab:module}
\end{table}
\begin{table}
\centering
\resizebox{1.0\linewidth}{!}{
\begin{tabular}{ccccccccc}
\hline
\multirow{2}{*}{\# GCN}  & \multirow{2}{*}{Adj} & \multirow{2}{*}{Pool} & \multirow{2}{*}{IIIT5k} & 
\multirow{2}{*}{SVT}& 
\multirow{2}{*}{IC15} & 
\multirow{2}{*}{CUTE}& \multirow{1}{*}{Params} &\multirow{1}{*}{Time}\\
&&&&&&& ($\times 10^{6}$) & (ms)  \\
\hline
2 & \{0,1\} & Graph   & 95.8  & 94.1  & 84.6 & 92.3 & 42.1 & 18.8\\
\hline
1 & \{0,1\} & Graph            & 94.3  & 92.9  & 82.5 & 90.8 & 39.5 &16.1\\
3 & \{0,1\} & Graph            & 96.0  & 94.0  & 84.8 & 92.6 & 44.9 & 22.5\\
\hline
2 & [0,1]  &Graph        & 95.9  & 94.2  & 84.9 & 92.4& 42.2 & 20.3\\  
2 & \{0,1\}  & Average        & 94.8  & 93.2  & 82.3 & 89.8& 38.1 & 15.7\\ 
\hline
\end{tabular}
}
\caption{ Ablation study of GTR. ``\#'' means  the number of  GCN layers in the first stage of GTR. ``Adj'' is the value type of adjacency matrix, $i.e.$, discrete value 0 or 1 and continuous value in [0,1], respectively. ``Average'' denotes  employing average pooling on graph feature rather than the graph pooling described in Section~\ref{subsec:gtr}.}
\label{tab:GTR}
\end{table}

\subsubsection{Influence of Different Settings in GTR}
Although the proposed GTR module has shown its effectiveness in improving the STR performance on multiple benchmarks, we would also like to analyze the influence of different settings in GTR. In this section, we evaluate the performance of GTR variants with respect to different numbers of GCN layers in the first stage, different value types of adjacency matrix, and different pooling strategies. As shown in Table~\ref{tab:GTR}, with the increase of the number of GCN layers, the recognition accuracy, the number of parameters, and inference time all increase as well. To achieve a trade-off between recognition accuracy and model complexity, we choose 2 layers as the default setting. Besides, we find that there is almost no performance gain when using continuous values in the adjacency matrix compared to discrete values, while the inference time increases by $7.98\%$. Therefore, we choose the discrete value as the default value type. We further compare the graph pooling with the average pooling in the stage of contextual reasoning. The results show that graph pooling outperforms average pooling significantly since it can capture the local-to-global dependency for reasoning. Therefore, we choose it as the default pooling strategy.

\begin{table}
\resizebox{1.0\linewidth}{!}{

\begin{tabular}{lcccccc}
\hline
 Fusion     &IIIT5k & SVT  & IC13 & SVTP & IC15 & CUTE \\ \hline
    Add & 94.8   & 93.2 & 95.0 & 84.9 & 83.3 & 90.8 \\ 
    Concat   & 95.0   & 93.4 & 95.4 & 85.1 & 83.4 & 91.3 \\ 
    D-fuse   & 95.8   & 94.1 &96.8 & 87.9 & 84.6 & 92.3 \\ \hline
\end{tabular}
}
\caption{Empirical study of the fusion strategy in S-GTR. ``Add'': element-wise addition. ``Concat'': Concatenation. ``D-fuse'': Dynamic fusion.}
\label{tab:fusion}
\end{table}

\subsubsection{Impact of Fusion Strategy}
We also investigate the impact of fusion strategy in S-GTR when fusing the linguistic context from LM and spatial context from GTR. In addition to the proposed dynamic fusion, we consider other two choices, $i.e.$, element-wise sum and concatenation. The results are reported in Table~\ref{tab:fusion}. As can be seen, while the concatenation fusion strategy performs better than element-wise addition, it still falls behind the proposed dynamic fusion strategy. We suspect that the benefit may come from the learnable fusion weights which are absent in the other two non-parametric cases.

\begin{figure}
	\centering
	\includegraphics[width=1.0\linewidth]{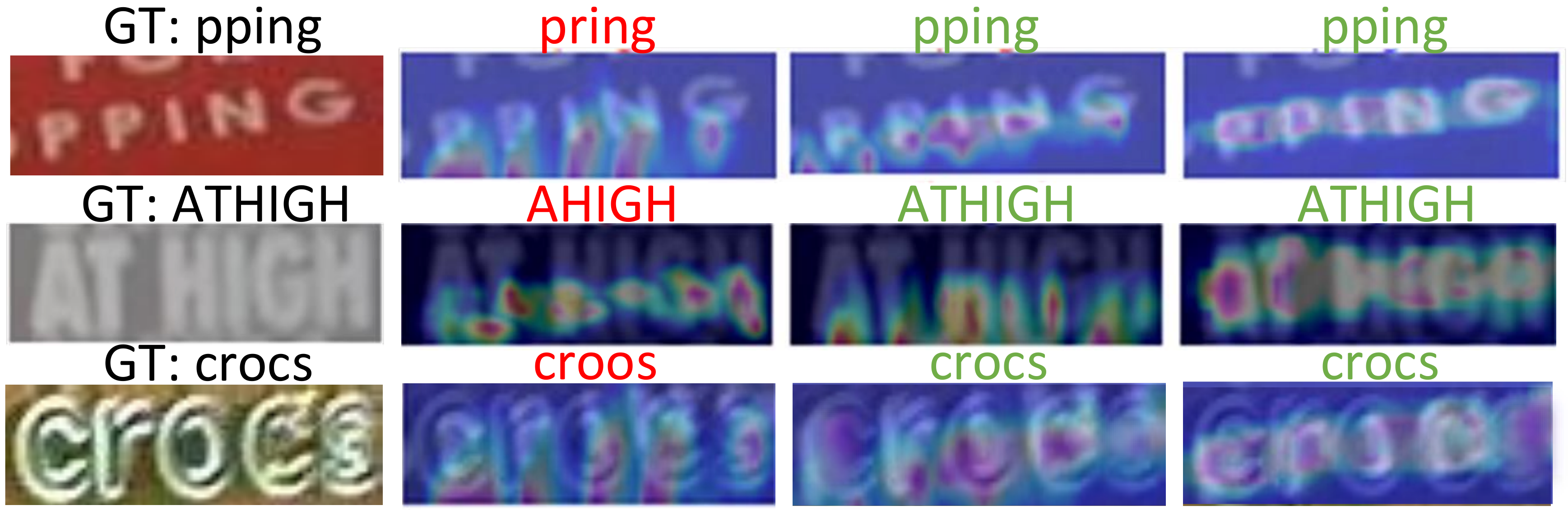}
	\caption{Visualization of feature maps from VR, GTR and S-GTR (from the second column to the last column).}
	\label{fig:att}
\end{figure}

\subsection{Further Visualization and Analysis}
\subsubsection{Visual Inspection Result}

\begin{figure}
	\centering
	\includegraphics[width=1.0\linewidth]{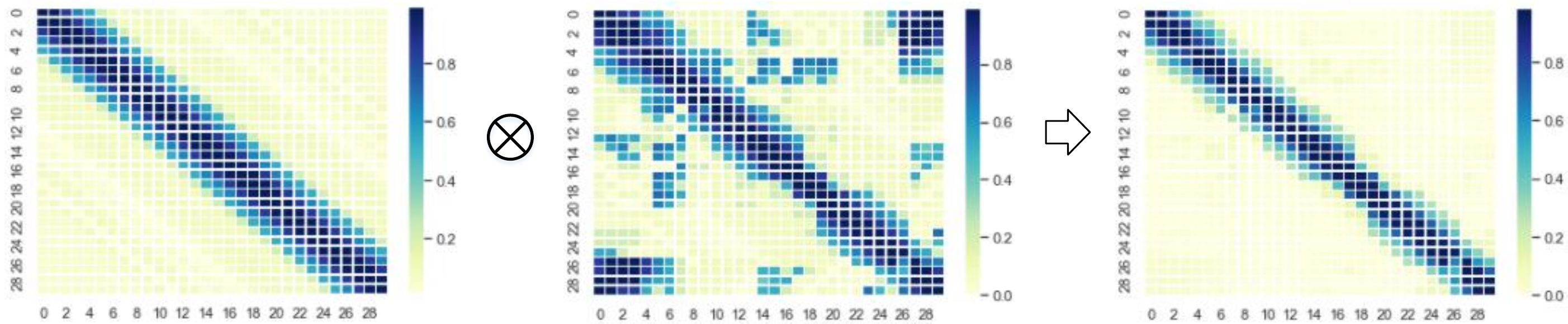}
	\caption{Visualization of a node similarity matrix, which is calculated according to Eq.~\eqref{eq:ps}.}
	\label{fig:visual11}
\end{figure}

For the qualitative analysis, we visualize the feature maps from the penultimate layer in VR, GTR and S-GTR. As shown in Figure~\ref{fig:att}, compared to the feature maps from VR, the feature maps from GTR are more strongly activated on the target characters owing to the textual reasoning of spatial context. Besides, the feature maps from S-GTR cover the target characters more precisely than GTR. These results imply that the S-GTR can learn more discriminative features by attending to the target characters and discard irrelevant information. In addition, we present the visualization of the node similarity matrix in Figure~\ref{fig:visual11} for better understanding the graph generation process.

\begin{table}
\footnotesize
\centering
\resizebox{1.0\linewidth}{!}{
\begin{tabular}{cccccccc}
\hline
Model & LM & IIIT5k & SVT  & IC13 & SVTP & IC15 & CUTE \\ 
\hline
VR & FastText  & 93.9 & 90.5 & 92.6 & 84.0 & 82.8 & 87.2 \\
S-GTR & FastText  & 94.8  & 92.9 & 95.1 & 85.3 & 84.0 & 90.6\\
\hline
VR  & BERT    & 94.3  & 92.0 & 94.0 & 85.6 & 83.8 & 90.8 \\ 
S-GTR & BERT  & 95.8  & 94.6 & 96.7 & 87.3 & 85.0 & 92.5 \\
\hline
\end{tabular}
}
\caption{Results of S-GTR with different language models. FastText and BERT are two pretrained language models.}
\label{tab:GTR-LM}
\end{table}

\begin{table}
\footnotesize
\centering
\resizebox{1.0\linewidth}{!}{
\begin{tabular}{cccccc}
\hline
&CRNN&ASTER & TextScanner &ABINet &S-GTR\\
\hline
Acc (\%) &59.2 & 57.4 & 64.1& 68.4& 72.2 \\
NED  &0.68 & 0.69& 0.75& 0.79  &0.82 \\
\hline
\end{tabular}
}
\caption{Results of different methods on MLT-17. "NED" is short for Normalized Edit Distance.}
\label{tab:chinese}
\end{table}

\subsubsection{Compatibility of GTR to Different LMs}
To further investigate the compatibility of GTR to LMs, we apply GTR in a basic VR model with two different LMs, $i.e.$, FastText \cite{bojanowski2017enriching} and BERT \cite{Devlin2019BERTPO}. As shown in Table ~\ref{tab:GTR-LM}, GTR contributes to consistent gains on both FastText and BERT Language Model. In addition, we also find that using a better LM together with GTR can further improve text recognition performance. 


\subsubsection{Chinese Scene Text Recognition}
Like English text recognition, the Chinese scene text recognition task offers an alternative way to assess the capability of STR models. The Chinese STR task is more challenging as it requires model to handle a larger vocabulary and more complex data associations. In addition to the recognition accuracy, we also report the Normalized Edit Distance (NED) of different methods following {\bt the ICDAR-2019 ReCTS \cite{zhang2019icdar}. As shown in Table~\ref{tab:chinese},} S-GTR outperforms other methods significantly on the multi-language dataset MLT-2017 \cite{nayef2017icdar2017}. 
It demonstrates that GTR is still very effective for textual reasoning of Chinese text materials.

\section{Conclusion}
In this paper, we propose the idea of performing textual reasoning based on visual semantics from a basic visual recognition (VR) model for the Scene Text Recognition (STR) task. We implement it as a graph-based textual reasoning (GTR) module, which can act as an easy-to-plug-in module in existing representative methods. It is shown to be very effective in improving STR performance while being complementary to the common practice, $i.e.$, language model-based linguistic context reasoning. Experimental results on six challenging STR benchmarks demonstrate that GTR can be plugged in different types of state-of-the-art STR models and improve their recognition performance further. GTR also shows good compatibility with different language models. Based on a simple segmentation-based VR model, we construct a simple S-GTR baseline for STR, which sets state-of-the-art on both English and Chinese regular and irregular text materials. We hope this work can provide a new perceptive to study textual reasoning in the STR task and inspire more follow-up work in the future, such as efficient design for spatial context-based reasoning as well as the way of effective fusion of different reasoning results.

\section*{Acknowledgements}
This work was supported in part by National Natural Science Foundation of China under Grants (No.62076186, No.61822113), and in part by Science and Technology Major Project of Hubei Province (Next-Generation AI Technologies) under Grant (No.2019AEA170). The numerical calculations in this paper have been done on the supercomputing system in the Supercomputing Center of Wuhan University.


\bibliography{aaai22}



\end{document}